\documentclass[11pt]{article}

\usepackage[utf8]{inputenc}
\usepackage[T1]{fontenc}
\usepackage[margin=1in]{geometry}
\usepackage{graphicx}
\usepackage{booktabs}
\usepackage{array}
\usepackage{caption}
\usepackage{amsmath}
\usepackage[authoryear,round]{natbib}
\usepackage[hidelinks]{hyperref}
\usepackage{float}
\usepackage{placeins}
\usepackage{lineno}
\graphicspath{{figures/}}

\title{\textbf{ 
From Gentlemen to Frontiermen: Masculine Formations in English-Language Fiction (1771--1930)}}
\author{Rong Wang\\ University of Tuebingen}
\date{}

\begin{document}
% \linenumbers (disabled for arXiv preprint)
\maketitle

\begin{abstract}
\noindent Masculinity in nineteenth-century fiction is not a single ideal but a
field of competing scripts. Drawing on 150 British and American canonical novels from the txtLAB Novel450 corpus, published between 1771 and 1930, this paper examines the changing relative prominence of competing models of masculine authority. To focus the analysis on masculine characterisation, the study extracts male-character-centred text windows by using coreference resolution to group names, nominal mentions, and pronouns into character-specific reference chains.
%The analysis focuses on passages centred on male characters, identified by using coreference resolution to group character names, nominal mentions, and pronouns into character-specific reference chains. 
It then fits an unsupervised structural topic model with publication year and author gender as topic-prevalence covariates. The model identifies six
distinct masculine formations: aristocratic-chivalric, Christian manhood,
gentlemanly respectability, country squire, professional-commercial, and
imperial/adventure. Across the corpus, formations tied to inherited rank and
sacred authority decline, while those organised around paid work and adventure
rise. The largest increase occurs not in professional-commercial breadwinning
but in imperial/adventure masculinity, particularly the frontier-wilderness
register. The trajectory points to a reallocation from inherited and sacred
status towards achieved, commercial, and expansionary forms of masculine
authority. Adventurous and commercial
formations are also more prevalent in novels by authors recorded as male. Because
these formations emerge without a seeded vocabulary yet align with categories established in
independent scholarship, the article offers a reproducible method for measuring
the reorganisation of gendered authority across the long nineteenth century.
\end{abstract}


\section{Introduction}

Masculinity has long been understood in gender studies as plural rather than singular \citep{connell1995masculinities,connell2005hegemonic}. Manhood is not a fixed essence but a field of competing arrangements: hegemonic and marginal, residual and emergent, organised through class, work, religion, and empire. For the long nineteenth century, historians and critics have traced this plurality through figures such as the gentleman, the father, the soldier, the clerk, and the man of God, each of whom inhabits a different script of manliness \citep{tosh1999mans,davidoff1987family,sussman1995victorian,adams1995dandies,mallett2015victorian}. Yet this scholarship has less often measured how the \emph{relative weight} of these scripts shifts across a century of fiction, or whether such shifts can be observed reproducibly rather than inferred from selected canonical examples. This article addresses that gap by asking a deliberately narrow question: when the unit of analysis is passages narrated about men, how does English-language fiction reweight competing masculine formations between 1771 and 1930?

To answer this question, we fit a structural topic model (STM) to 150 English-language novels published between 1771 and 1930, with publication year and author gender modelled as topic-prevalence covariates \citep{roberts2019stm}. Fiction is a useful archive for this question because it does not merely represent different kinds of men; it also assigns narrative space and value to different forms of masculine authority. The methodological challenge is to measure that allocation without modelling everything else a novel contains. A topic model of whole novels would capture women, children, settings, plot machinery, and general narrative atmosphere alongside passages about men. We therefore model only passages narrated \emph{about men}, identified through character-level coreference rather than through a masculine word list.Character mentions are resolved into entities, assigned gender, and used to retain windows in which male characters predominate. On this male-referent corpus we fit an
\emph{unsupervised} model, seeding no masculine vocabulary and fixing no topic in
advance, so the formations are discovered from word co-occurrence and labelled
afterward.

The model reveals a clear historical pattern. Across the long nineteenth century, formations of inherited and sacred authority give way to formations organised around paid work and adventure (Section~\ref{sec:traj}). It shows how a major literary corpus reallocates narrative space among scripts of authority, work, religion, class, and empire. The article therefore offers both a long-run account of that shift and a method for making computational gender analysis refer to male figures rather than to gendered vocabulary alone.

\section{Related work}

This article builds on topic modelling and distant reading. In literary history, and in the cultural sciences more broadly, topic modelling has offered a way to infer thematic structure without hand-coding texts \citep{tangherlini2013trawling,goldstone2014quiet,mohr2013introduction}, one strand
of the broader distant-reading programme \citep{moretti2013distant}.
\citet{jockers2013significant} tied topic prevalence to author gender, nationality,
and date. We adopt STM \citep{roberts2014structural,roberts2019stm} rather than latent
Dirichlet allocation \citep{blei2003lda}, because STM allows publication year and author gender to enter the model directly as prevalence covariates. At the same time, topic models require interpretive caution. Their outputs are artefacts to be interpreted, not conclusions in themselves \citep{buurma2015fictionality}. Better-fitting models can be less interpretable \citep{chang2009reading,mimno2011optimizing}, automated coherence metrics can favour models that human readers reject \citep{hoyle2021automated}, and some computational literary findings have proved statistically fragile \citep{da2019computational}. We therefore keep interpretation in the loop and
check the model against external evidence.

This study extends computational research on gender representation. Work in this area has traced changing gender stereotypes through diachronic word embeddings \citep{garg2018word} and measured the power and agency that texts project onto male and female characters \citep{sap2017connotation}. We share their aim of measuring gendered representation at
scale while keeping the object literary and historical.
For the present study, the crucial precedent is character-level modelling.
\citet{underwood2018transformation} attribute words to characters to track the
changing marking of gendered characters, and modern coreference pipelines such as
BookNLP \citep{bamman2014bayesian} make character-level resolution increasingly practical for literary fiction. We use this machinery to define the object of analysis: passages narrated about men, which allows the model to operate on a male-referent corpus rather than on whole novels or on a preselected list of masculine terms.

The resulting formations are interpreted against historical scholarship on masculinity. \citet{connell1995masculinities} makes plurality, hierarchy, and class central to the analysis of masculinity; \citet{tosh1999mans} and \citet{davidoff1987family} trace middle-class manhood through work, family, and religion; and \citet{sussman1995victorian}, \citet{adams1995dandies}, \citet{solinger2012becoming}, and \citet{mallett2015victorian} examine gentility, asceticism, muscular Christianity, and masculine self-fashioning in literary and cultural history. These studies provide the external reference points against which we judge the modelled formations: chivalric, Christian, gentlemanly, commercial, and frontier masculinities.

\section{Data and method}

\subsection{Corpus}
The corpus is the English-language subset of Andrew Piper's txtLAB Novel450 dataset
\citep{piper2016txtlab}: 150 British and American canonical novels published between 1771 and
1930. Novel450 is a curated collection of canonical literary fictions rather than an opportunistic archive, a single-author corpus, or a narrowly genre-specific sample. The subset includes 108 British and 42 American novels, with 81 works by authors recorded as male and 69 by authors recorded as female. It spans five literary-historical periods (Georgian, Early Victorian, Late Victorian, Edwardian, and Early Modernist) and contains 107 third-person and 43 first-person narratives across 98 authors. The metadata record publication date, decade, author, title, nationality, author gender, narrative person, length, and language. The corpus therefore provides a controlled long-run sample for examining changes in fictional masculinity across period, national tradition, narration, and author gender.  %We therefore treat the corpus as a curated literary-historical sample of canonical English-language fiction, not as a complete population of nineteenth-century novels.

\subsection{Selecting the male-referent corpus}
\label{sec:maleref}
Because a topic model measures whatever vocabulary a document contains, modelling
whole novels would conflate masculine discourse with everything else a novel
narrates. We therefore restrict the model to passages narrated about men, which we
identify with character-level coreference rather than a word list. No
masculinity keyword, role list, or dictionary enters this selection; only the
coreference-based balance of male and female characters.

Each novel is
processed with BookNLP, which clusters every character mention, including proper
names, common-noun descriptions, and pronouns, by coreference. It also infers each
character's gender, in the tradition of computational character analysis for English fiction
\citep{bamman2014bayesian,underwood2018transformation}. We then divide the novel
into non-overlapping, sentence-aware windows of about 500 words. This is a standard
chunk size for topic-modelling fiction, since chunks below roughly 250 words yield
incoherent topics and 500 to 1000 words is common practice
\citep{jockers2013macroanalysis}. We then count, for each window, the mentions resolved
to male versus female characters. A window enters the corpus if its male mentions
exceed its female mentions, with at least two gendered mentions present; the rest
are set aside. Of 35{,}434 windows, 22{,}197 (63\%) are male-referent and form the
modelling corpus, while 12{,}831 are female-dominant and 406 are tied or below the
gendered-mention threshold. All 150 novels are represented. 

As a check that this
balance is not corrupted by the known difficulty of coreference on archaic prose,
the retained windows are independently male-dominant by raw third-person pronouns,
a signal coreference does not reduce to, with a male-pronoun share of $0.75$ to $0.77$
in every period from Georgian to Early Modernist.This check suggests that the selection procedure does not suffer a visible period-specific deterioration on the earliest prose.

This selection trades recall for precision. It makes the modelled corpus predominantly male-referent rather than merely masculine in vocabulary. That condition provides the basis for interpreting the resulting topics as masculine formations and reduces the risk of mistaking female-centred topics of home and feeling for masculine ones. But it sets aside the roughly one-third of passages in
which women dominate, and with them much of masculinity as it is constituted \emph{in
relation to} women: the father among his children, the suitor, the husband. We
therefore measure masculinity in male-centred narrative space rather than the whole
discourse of manhood. We keep this boundary explicit and revisit it in the conclusion.

\subsection{Preprocessing and estimation}

Because preprocessing decisions can materially shape the output of unsupervised
text models \citep{denny2018text}, ours is deliberately conservative and free of seed vocabulary.
Tokens are lowercased and normalised, stopwords are removed, and very rare or
ubiquitous terms are filtered (terms are retained when they appear in at least 25
documents and in no more than 65\% of documents); no masculine vocabulary is seeded or protected.
Character names are removed without an external name list, by exploiting a
statistical property of proper nouns. A personal name recurs heavily within the one
novel it belongs to but is largely absent from the rest of the corpus, so it is highly
document-specific in a way ordinary vocabulary is not. We therefore remove any term that
occurs in no more than 10\% of the documents yet reaches at least 20 occurrences in a
single document. This removes the
bulk of character and place names, along with novel-specific coinages and residual
transcription artefacts, while leaving vocabulary shared across the corpus intact.
Removing names by this data-driven criterion, rather than by a name gazetteer, has two
advantages here. It keeps the pipeline free of any external word list, consistent with
the study's dictionary-free design, and it does not rely on capitalisation, a cue that
lowercasing has already discarded and that a gazetteer would need. The step leaves a
final vocabulary of 13{,}195 terms.

We estimate
the model with the \texttt{stm} package \citep{roberts2019stm}, which implements the
structural topic model \citep{roberts2014structural,roberts2016model}, using
Spectral initialisation based on the anchor-word algorithm
\citep{arora2013practical} and the prevalence formula \texttt{year + author\_gender}.
Publication year and author gender, coded as recorded in the metadata,enter the model as predictors of topic prevalence. The random seed is fixed at 42 for reproducibility, and estimation runs for a maximum
of 75 EM iterations.
Topics are labelled after estimation by reading their highest-probability and
frequency-exclusivity (FREX) words \citep{bischof2012summarizing}. Uncertainty on
period prevalences is quantified by a novel-level bootstrap that resamples whole
novels within each period (4{,}000 replicates), so the intervals reflect the number
of independent novels rather than the much larger passage count.

\subsection{Choosing the number of topics}
\label{sec:K}
\begin{figure}[H]
\centering
\includegraphics[width=0.9\textwidth]{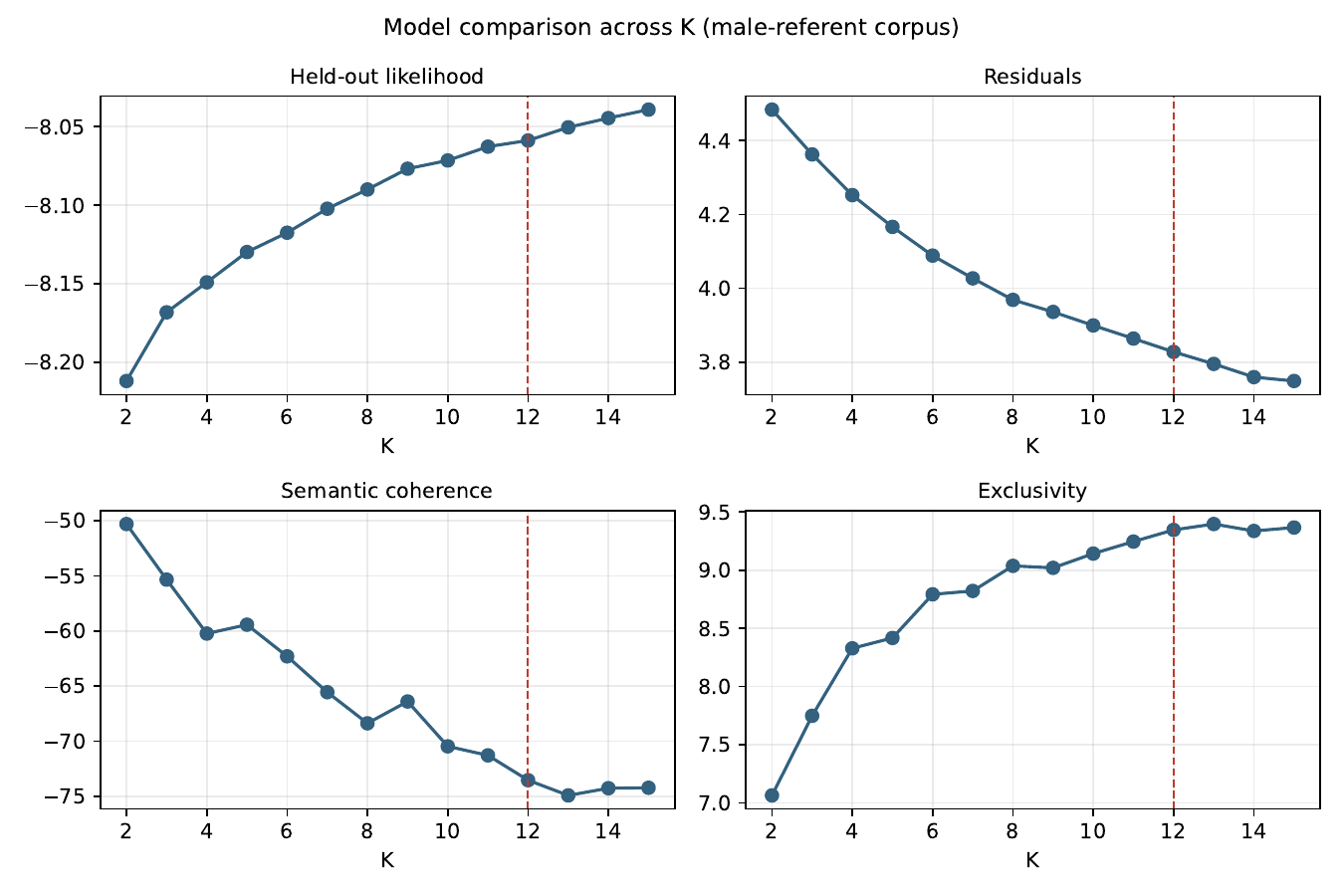}
\caption{Model comparison across $K=2$ to $15$ on the male-referent corpus, on four
standard diagnostics. Held-out likelihood, residuals, and exclusivity improve
monotonically with $K$, whereas semantic coherence is highest at small $K$, with local
maxima at $K=5$ and $K=9$. No single $K$ optimises all four criteria, so the reported
model, $K=12$ (dashed line), is chosen on the interpretive ground given in
Section~\ref{sec:K} rather than at a diagnostic optimum.}
\label{fig:modelcomp}
\end{figure}

We fit models for $K=2$ to $15$ and inspected the standard diagnostics
(Table~\ref{tab:modelcomp}, Fig.~\ref{fig:modelcomp}). As is usual, held-out
likelihood \citep{wallach2009evaluation}, residuals, and exclusivity improve
monotonically with $K$, whereas semantic coherence \citep{mimno2011optimizing} is
highest at small $K$, with local maxima at $K=5$ and
$K=9$, and declines thereafter. The diagnostics do not identify a single optimum. We
report $K=12$ on an interpretive criterion, because it is the smallest model in which the
theoretically distinct formations that the scholarship recognises emerge as
\emph{separate} topics rather than fused. At $K\le10$ the aristocratic-chivalric and
Christian formations share a single topic and the country-squire formation does not
appear. The search range extends beyond the reported model, so $K=12$ is not an
upper-bound choice. The substantive trajectories reported below are not dependent on this choice.

The rise-and-decline directions reported below hold at every $K$ from $7$ to $12$,
the coherence-optimal $K=9$ included. At each of these sizes the
professional-commercial and imperial/adventure formations rise, while the
aristocratic-chivalric, gentlemanly, and country-squire formations decline.
Increasing $K$ beyond this range mainly splits existing formations into finer ones
rather than reversing any direction. The coherence at $K=12$, though lower than at
$K=9$, reflects the usual cost of that finer granularity rather than incoherent
topics.

\begin{table}[htbp]
\centering
\caption{Model comparison across $K$ (male-referent corpus). Held-out likelihood,
residuals, and exclusivity improve with $K$; semantic coherence is best at small
$K$ (local maxima at $K=5,9$). No single optimum; $K=12$ chosen for interpretability}
%(Section~\ref{sec:K}).}
\label{tab:modelcomp}
\small
\begin{tabular}{r r r r r}
\toprule
$K$ & Held-out & Residual & Sem.\ coherence & Exclusivity \\
\midrule
2  & $-8.212$ & 4.484 & $-50.30$ & 7.06 \\
5  & $-8.130$ & 4.166 & $-59.44$ & 8.42 \\
7  & $-8.102$ & 4.027 & $-65.56$ & 8.82 \\
9  & $-8.077$ & 3.936 & $-66.40$ & 9.02 \\
10 & $-8.072$ & 3.900 & $-70.48$ & 9.14 \\
\textbf{12} & $\mathbf{-8.059}$ & \textbf{3.828} & $-73.55$ & \textbf{9.35} \\
15 & $-8.039$ & 3.749 & $-74.23$ & 9.37 \\
\bottomrule
\end{tabular}
\end{table}

\section{Results}

\subsection{The masculine formations}
Estimated without any seeded vocabulary, the model returns topics that closely correspond to recognised masculine formations (Table~\ref{tab:topics}). We label each topic from
its most probable words and representative passages rather than from
frequency-exclusivity (FREX) words alone. Because FREX rewards distinctiveness, its
top entries over-represent forms of polite address (\emph{sir}, \emph{lordship},
\emph{madam}) and residual character names (\emph{slope}, \emph{fang}, \emph{buck})
that escaped name removal, which can make a substantive formation look like a mere
register. The probability words better capture the dominant register: the
gentlemanly topic foregrounds social relations and correspondence (\emph{lady},
\emph{sir}, \emph{friend}, \emph{dear}, \emph{letter}), while the frontier topic
foregrounds physical action (\emph{hand}, \emph{head}, \emph{face}, \emph{men},
\emph{stood}). The remaining
topics are narration registers, such as drawing-room sociability, emotional
interiority, colloquial dialogue, and Gothic setting. Because they are not scripts
of manhood, we treat them separately. 

Six formations are central to the argument, and their fuller
vocabularies (the leading frequency-exclusivity words of each,
Table~\ref{tab:topics}) support the labels. An \emph{aristocratic-chivalric} topic
of rank and martial honour gathers
the medieval and dynastic lexicon (\emph{knight, prince, sword, duke, baron, king,
castle}) with \emph{friar, saxon, highness, warrior}, and \emph{gallant} placing it
in the chivalric revival and the historical romance of Scott and his heirs
\citep{girouard1981camelot}. A \emph{Christian manhood}
topic of sacred authority and moral manhood runs on \emph{god, sin, soul, prayer,
priest, heaven, christ, holy, faith}, and \emph{mercy}, the muscular Christianity of
the Victorian pulpit \citep{vance1985sinews}. \emph{Gentlemanly
respectability}, the masculinity of polite urban society, is dominated by titles
(\emph{lord, lordship, lady, madam, miss}) and by the vocabulary of the marriage
plot (\emph{letter, proposal, consent, pardon, dear}), the courtship and
correspondence of Burney and Austen. The landed \emph{country squire} gathers the
rural estate (\emph{squire, horse, coach, landlord, parish, stable, pony,
riding}) and the wider village world of \emph{farmer, doctor, folk, inn}, and
\emph{brandy}. The \emph{professional-commercial breadwinner} runs on money and
modern infrastructure (\emph{dollar, pound, cent, money, sell, buy, trade, loan,
wage, office, railroad, railway, wheat}), the world of finance, commerce, and the
market. Finally, an \emph{imperial/adventure} masculinity, the man tested against
sea, frontier, and empire, combines a maritime register (\emph{ship, sea, sailor,
whale, deck, shore, vessel, crew}) and a land/frontier register (\emph{wolf, dogs,
pack, jungle, cub, sergeant}), the wild of Jack London, Kipling, and Crane.

The model separates gentility into two distinct formations rather than one. Despite a
shared social milieu, gentlemanly respectability and the country squire are the
polite gentleman of urban society and the landed gentleman of the rural estate,
respectively: the first is the masculinity of manners, courtship, and formal address
that animates Burney and Austen, the second the masculinity of horses, the parish,
and the field that animates novels such as \emph{Black Beauty} and \emph{Castle
Rackrent}. They correlate near zero across novels and share almost no vocabulary,
so we keep them separate. The imperial/adventure formation is the mirror case. The
model again returns two topics, the maritime and the land/frontier registers described
above, but the line between them is one of setting rather than of masculinity. Both
narrate the same script, the man tested against a wild and dangerous world beyond
settled society, and both belong to the same literature of adventure and empire. What
changes between them is only where the test is staged, at sea or on the frontier. To
count them as two formations would raise a difference of scenery to a difference of
manhood, so we read them as a single formation with two settings. This yields
six masculine formations in total.

\subsection{The historical reweighting of masculinity}
\label{sec:traj}
\begin{figure}[H]
\centering
\includegraphics[width=\textwidth]{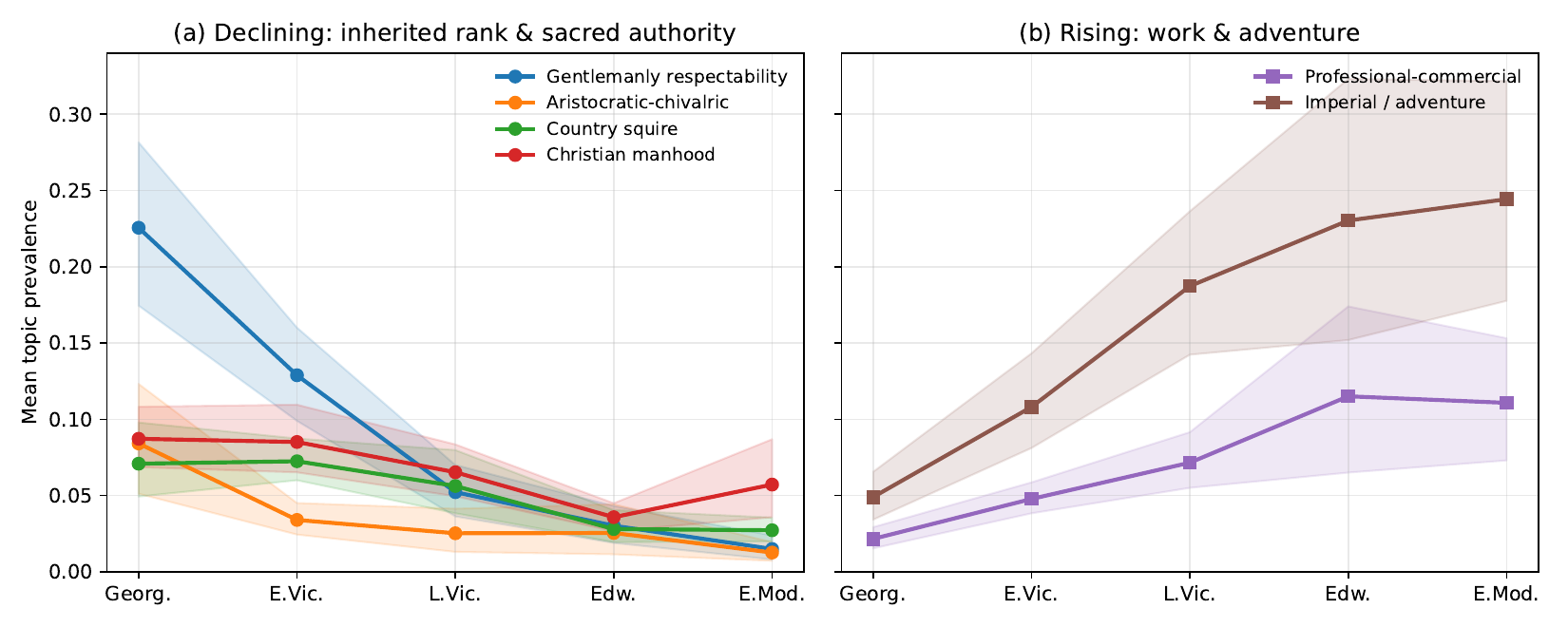}
\caption{Period trajectories of the six masculine formations (male-referent
corpus, $K=12$; imperial/adventure combines the model's sea and frontier topics). Periods are abbreviated Georg. (Georgian), E.Vic. (Early Victorian), L.Vic. (Late Victorian), Edw. (Edwardian), and E.Mod. (Early Modernist).
Shaded bands are 95\% novel-level bootstrap intervals. The formations of inherited
rank and sacred authority decline (the gentlemanly, aristocratic-chivalric, and
country-squire declines are significant; the Christian-manhood decline is not);
the professional-commercial and imperial/adventure formations rise
significantly.}
\label{fig:traj}
\end{figure}

Figure~\ref{fig:traj} and Table~\ref{tab:topics} give the trajectories with 95\%
novel-level bootstrap intervals. The formations of inherited and sacred status
recede, while those of paid work and adventure rise. Gentlemanly respectability falls most sharply, from the most prevalent
masculine formation in the Georgian novel to near-absence by the modernist period
($0.226 \rightarrow 0.015$; $-0.211$, 95\% CI $[-0.267, -0.158]$, $p<0.001$).
Against this retreat, the imperial/adventure formation rises most steeply
($0.049 \rightarrow 0.244$; $+0.195$ $[+0.126, +0.274]$, $p<0.001$). Within it the
rise is carried by the land/frontier-wilderness register, which climbs from
near-absence ($0.014$) to the most prevalent masculine register of the modernist
period ($0.204$). Its maritime/seafaring counterpart is comparatively flat and, taken alone,
$K$-sensitive and carried by a few canonical sea-novels (\emph{Moby-Dick},
\emph{Treasure Island}, Cooper's romances); the adventurous rise thus reflects the
turn to the frontier and the wild rather than to the sea.

\begin{table}[H]
\centering
\caption{The six masculine formations recovered from the male-referent corpus
($K=12$): trajectory (Georgian $\rightarrow$ Early Modernist prevalence), the change
$\Delta$ with 95\% novel-level bootstrap interval, mean prevalence by author gender
(M, F), and leading frequency-exclusivity (FREX) words. Residual character names and
lemmatised forms, which dominate the raw exclusivity lists, are excluded from the
words shown. The imperial/adventure formation groups the model's sea/maritime and
land/frontier topics, which differ only in setting; labels are assigned post hoc.}
\label{tab:topics}
\footnotesize
\begin{tabular}{@{}>{\raggedright\arraybackslash}p{2.3cm} c >{\raggedright\arraybackslash}p{2.8cm} c c >{\raggedright\arraybackslash}p{4.3cm}@{}}
\toprule
Formation & Geo.$\rightarrow$Mod. & $\Delta$ (95\% CI) & M & F & Leading FREX words \\
\midrule
\multicolumn{6}{l}{\emph{Declining}}\\
Gentlemanly respectability & $0.226\!\rightarrow\!0.015$ & $-0.211$ \mbox{$[-0.267,\,-0.158]$} & 0.067 & 0.168 & madam, lordship, miss, mrs, lady, lord, pardon, letter, proposal, consent, dear, assure, uncle, bishop \\
\addlinespace
Aristocratic-chivalric & $0.084\!\rightarrow\!0.013$ & $-0.071$ \mbox{$[-0.111,\,-0.038]$} & 0.060 & 0.022 & knight, prince, sword, duke, baron, king, castle, friar, saxon, warrior, highness, royal, military, gallant \\
\addlinespace
Country squire (landed) & $0.071\!\rightarrow\!0.027$ & $-0.044$ \mbox{$[-0.071,\,-0.019]$} & 0.063 & 0.053 & squire, horse, ride, coach, landlord, pony, parish, stable, riding, farmer, doctor, folk, inn, brandy \\
\addlinespace
Christian manhood & $0.087\!\rightarrow\!0.057$ & $-0.030$ \mbox{$[-0.061,\,+0.005]$} & 0.073 & 0.071 & god, sin, soul, prayer, priest, heaven, christ, holy, faith, mercy, divine, blessed, angel, christian \\
\midrule
\multicolumn{6}{l}{\emph{Rising}}\\
Professional-commercial & $0.022\!\rightarrow\!0.111$ & $+0.089$ \mbox{$[+0.051,\,+0.132]$} & 0.081 & 0.037 & money, dollar, pound, cent, price, sell, sold, buy, loan, trade, wage, office, railroad, railway, wheat \\
\addlinespace
Imperial / adventure & $0.049\!\rightarrow\!0.244$ & $+0.195$ \mbox{$[+0.126,\,+0.274]$} & 0.179 & 0.093 & ship, sea, sailor, whale, deck, shore, vessel, crew; wolf, dogs, pack, jungle, cub, sergeant \\
\bottomrule
\end{tabular}
\end{table}

The other formations move in these same two directions but less dramatically. The
aristocratic-chivalric and country-squire formations also decline significantly,
and the professional-commercial breadwinner rises significantly
($0.022 \rightarrow 0.111$, $p<0.001$); Christian manhood drifts downward but not
significantly ($p=0.094$), so we do not count it among the robust trends. Their full
changes and 95\% intervals are given in Table~\ref{tab:topics}. In sum, the corpus
moves from the knight, the gentleman of rank, the country squire, and the man of God
towards the clerk, the breadwinner, and the man of the frontier. This is a
reweighting rather than a clean replacement: the older formations persist at lower
prevalence rather than vanishing.

These trajectories are not driven wholly by a few canonical texts. Removing, for
each formation, the five novels that load on it most heavily leaves the directions
intact: the frontier rise falls only from $+0.190$ to $+0.163$, the
professional-commercial rise from $+0.089$ to $+0.078$, and the gentlemanly decline
from $-0.211$ to $-0.162$, while the aristocratic-chivalric decline weakens but
persists, from $-0.071$ to $-0.033$. As with any trajectory over a 150-novel canon,
these movements are shaped by which genres fall in which period: courtship fiction
early, business and adventure fiction late. Yet they survive the removal of their
most influential novels. They are broader corpus patterns, even though genre
composition still shapes their size.

\subsection{Formations by author gender}

\begin{figure}[H]
\centering
\includegraphics[width=0.85\textwidth]{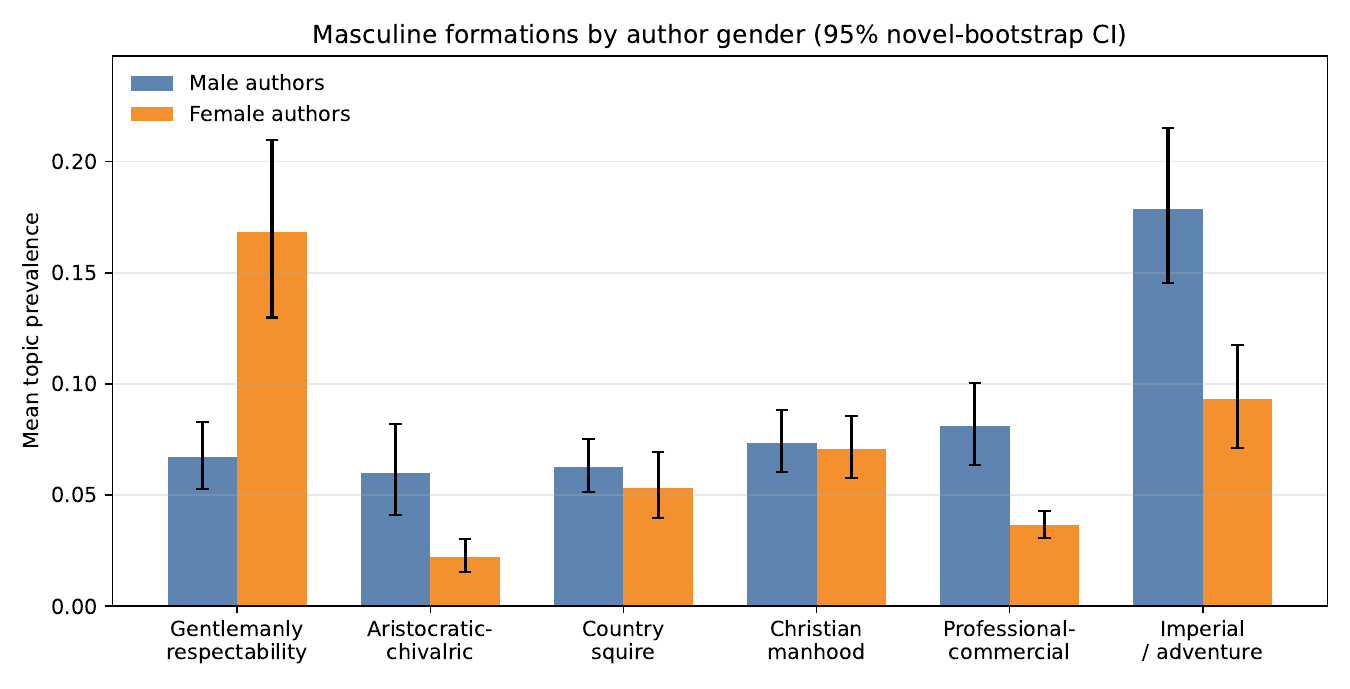}
\caption{Mean prevalence of the six masculine formations in novels by authors recorded
as male versus female, with 95\% novel-level bootstrap intervals. The
imperial/adventure, aristocratic-chivalric, and professional-commercial formations are
more prevalent in male-authored novels, gentlemanly respectability is more prevalent in
female-authored novels, and Christian manhood is roughly even between the two groups.
Because author gender enters the model only as a covariate, these are corpus-level
associations rather than claims about individual authors.}
\label{fig:authgender}
\end{figure}

Author gender is associated with the formations in historically interpretable ways
(Fig.~\ref{fig:authgender}). In the metadata's binary
author-gender field, adventurous and commercial formations are more prevalent in
novels by authors recorded as male: imperial/adventure ($0.179$ male vs.\
$0.093$ female), aristocratic-chivalric ($0.060$ vs.\ $0.022$), and
professional-commercial ($0.081$ vs.\ $0.037$). Gentlemanly respectability, by
contrast, is more prevalent in novels by authors recorded as female ($0.168$
vs.\ $0.067$), while Christian manhood is nearly even between the two groups.
Because author gender enters only as a prevalence covariate, these differences should be read as corpus-level associations rather than as claims about authorial intent.

\subsection{Nationality as a robustness check}
\label{sec:nation}
The corpus mixes British and American fiction unevenly across the period, and it
Americanises over time: the American share climbs from 17\% of Georgian novels to
28\% of Late Victorian and a peak of 67\% of Edwardian novels, then eases to 47\% of
Early Modernist ones.
Each author is coded British or American, with Scottish, Welsh, Irish, and
British-career authors grouped as British, and U.S. authors and immigrants of
mainly American career as American. Because the rising formations are prominent
in American naturalist and adventure fiction, this composition shift could in
principle produce the aggregate trend. The question, then, is whether the
same directions remain visible when British and American novels are examined
separately.

\begin{figure}[H]
\centering
\includegraphics[width=0.9\textwidth]{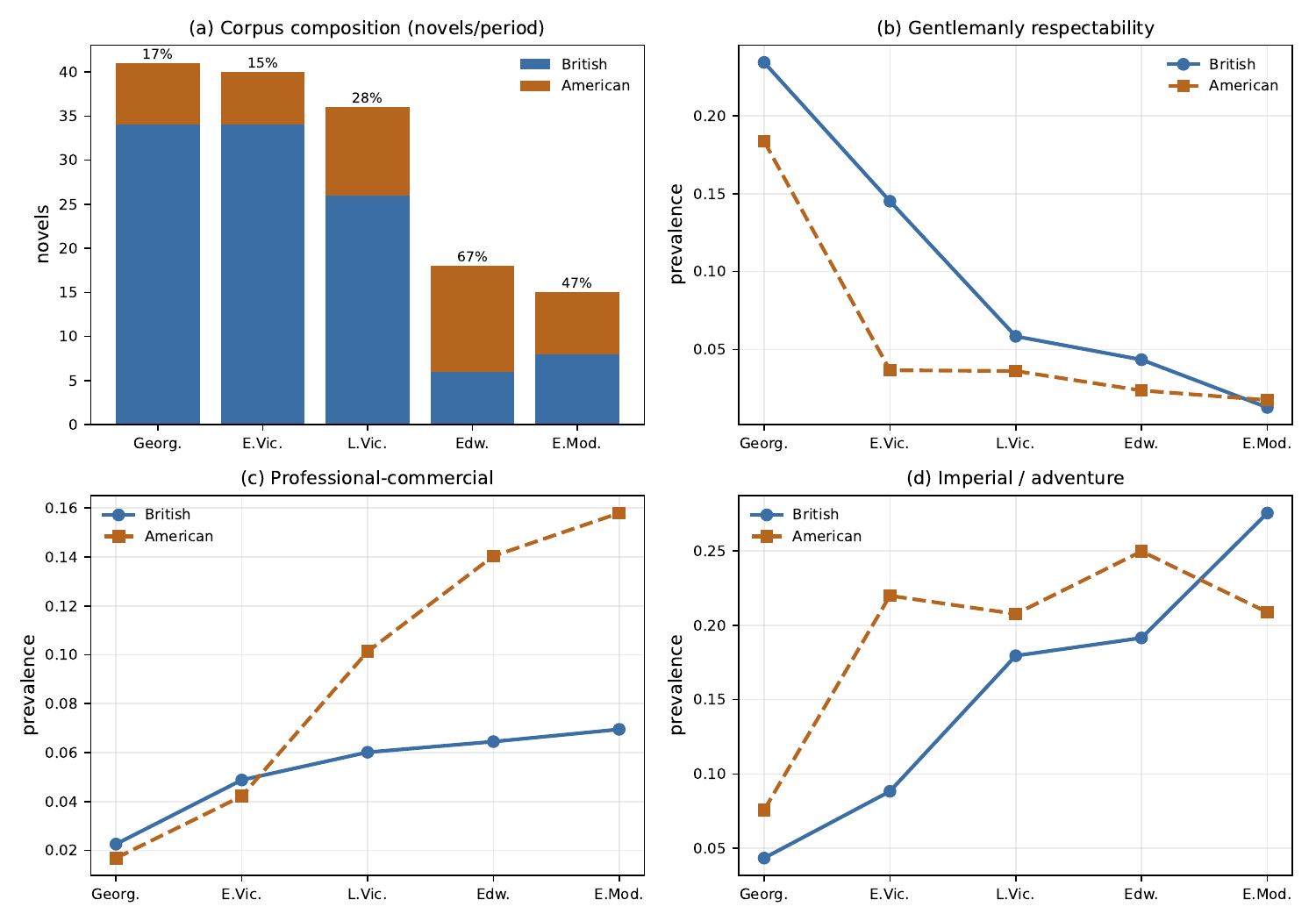}
\caption{Nationality and corpus composition. Periods are abbreviated Georg. (Georgian), E.Vic. (Early Victorian), L.Vic. (Late Victorian), Edw. (Edwardian), and E.Mod. (Early Modernist). (a) Corpus composition by period
(British vs.\ American novels; American share annotated), showing the late-century
Americanisation and the small late British samples. (b) to (d) Within-nation
period trajectories for three formations: gentlemanly respectability declines in
both nations, the imperial/adventure formation rises in both, and the
professional-commercial rise is led by American fiction (the British trajectory
is nearly flat).}
\label{fig:nation}
\end{figure}

The within-nation split argues against a purely compositional reading
(Fig.~\ref{fig:nation}). Nationality does not enter the model. When we split the fitted
topic proportions by it as a post-hoc robustness check, the central directions
hold within each nation. Gentlemanly respectability declines in both British
($0.234 \rightarrow 0.013$) and American ($0.184 \rightarrow 0.017$) fiction, and
the imperial/adventure formation rises in both (British $0.043 \rightarrow 0.275$;
American $0.076 \rightarrow 0.209$). The one nation-specific trend is the
professional-commercial rise: strong among American authors
($0.017 \rightarrow 0.158$) but weak among British ones ($0.023 \rightarrow 0.069$).
The breadwinner's ascent is therefore driven mainly by American business-oriented fiction in the corpus. 
The broader status reallocation is not an artefact of the British-to-American shift,
although that shift amplifies the aggregate professional-commercial trend.

\section{Discussion}

The results point to a reallocation of fictional masculine authority. Across the
long nineteenth century the formations of inherited and sacred manhood recede, while
those organised around paid work and adventure advance. The professional-commercial
breadwinner rises, but the imperial/adventure man rises fastest of all. That the model resolves gentility into two formations, the polite
gentleman of urban society and the landed gentleman of the rural estate, sharpens
the picture, since both retreat as the century turns s work and the frontier. 
The corpus does not simply exchange one ideal of manhood for another. It redistributes
narrative attention from authority a man inherits towards authority he must earn or
seize.

The pattern corroborates and refines the historiography. \citet{tosh1999mans} and
\citet{davidoff1987family} describe the consolidation of bourgeois work-and-family
manhood, and \citet{connell1995masculinities} makes class central to plural
masculinities. For the American case, where the rising formations concentrate,
\citet{kimmel1996manhood} traces the ascent of the self-made man and
\citet{bederman1995manliness} the turn-of-the-century shift from restrained moral
manliness towards an aggressive, ``primitive'' masculinity. Our trajectories register these movements at corpus scale as the American-led
professional rise and the climb of the frontier formation, and date them across five
literary-historical periods rather than inferring them from selected texts.

Beyond confirming these accounts, the model locates the steepest rise of the period,
and it belongs to the imperial/adventure formation, not to the bourgeois professional
that the historiography treats as the emblem of modern manhood. Within that formation
the gain is concentrated in the frontier and the wild, the masculinity of the trail
that American and late-century adventure fiction made central. This complicates the
familiar story of modernising manhood, in which the disciplined bourgeois replaces the
man of rank. Nor is that rise a simple democratisation. As inherited rank loses its hold, the registers that replace it,
frontier and martial, carry their own hierarchies of violence, race, and empire. The
reallocation is better read as a change in the terms of masculine authority than as
its levelling.

Several validation checks support this reading. The first is convergent. An
unsupervised model, given no list of masculine words, reproduces the chivalric,
Christian, gentlemanly, commercial, and frontier masculinities of independent
scholarship and recovers a periodisation those scholars established by close
reading. The second is referential. Because the corpus is defined by
character-level coreference, the formations describe passages narrated about men
rather than vocabulary that merely sounds masculine. The third is robustness. The
principal directions survive alternative topic numbers and leave-top-novel tests.
These checks do not make the labels automatic. The labels remain interpretive, and
period effects partly reflect genre and corpus composition. They do, however, make the
historical pattern a property of the corpus, rather than an artefact of our labelling
or a restatement of existing scholarship.

\section{Conclusion}

By modelling male-referent passages selected through character-level coreference, this article recovers six masculine formations across 150 English-language novels and traces their historical reweighting. The pattern is a reallocation of
fictional manhood. Formations of inherited rank and sacred authority (the gentleman,
the knight, the squire, the clergyman) lose ground to those of paid work and the
frontier (the clerk, the breadwinner, the frontiersman).

% %The scope of this claim is deliberately specific. The analysis measures
% masculinity in male-centred narrative space rather than the whole discourse of
% manhood. This restriction strengthens the design because the model does not infer
% masculinity from whole-novel themes or from vocabulary that merely appears
% masculine. Instead, it works from passages dominated by male characters. While it reduces the risk of mistaking passages narrated primarily about
% women for masculine formations, it sets
% aside some relational forms of masculinity, including the father, suitor, and
% husband. The selection procedure is supported by two
% checks. The retained windows are independently male-dominant by raw pronouns in
% every period, and the main trajectories survive the reported robustness checks
% across topic number and influential novels.
By selecting passages in which male characters predominate, the analysis avoids inferring masculinity from whole-novel themes or from vocabulary that merely appears masculine. It reduces the risk of treating passages centred on women, domestic feeling, or general social life as masculine formations. The trade-off is that some relational forms of masculinity, including fatherhood, courtship, husbandhood, and male dependence on women and children, may be less visible when they appear mainly in female-dominant passages. The results should therefore be read not as an exhaustive account of manhood in the novel, but as an account of masculinity in male-centred narrative space. This boundary is supported, though not erased, by the validation checks: the retained windows are independently male-dominant by raw pronouns in every period, and the main trajectories survive robustness tests across topic number and influential novels.

The paper makes both a historical and a methodological contribution. Historically,
the paper shows that English-language fiction reallocates masculine authority
from inherited and sacred status towards work, commerce, and frontier adventure.
Methodologically, it shows how character-level text selection can make distant
reading more precise. Because the formations emerge without a seeded vocabulary yet reproduce categories known from independent scholarship, the resulting trajectory offers a reproducible account
of how fiction reorganised masculine authority between 1771 and 1930.

\section*{Data availability}
The corpus is Andrew Piper's txtLAB Novel450 dataset \citep{piper2016txtlab}. 
The anonymised supplementary package includes the scripts for BookNLP
character-resolution processing, male-referent passage selection, preprocessing,
STM estimation, reporting, and figure generation. The reported model uses BookNLP's
English small model, \texttt{stm} 1.3.8, R 4.5.0, Python 3.13.2, seed 42,
\texttt{max.em.its = 75}, $K=12$, and the STM prevalence formula
\texttt{year + author\_gender}. The package will be deposited in a persistent
archive upon acceptance.

\section*{Ethics statement}
Ethical approval was not required: the study analyses published literary texts and
involves no human participants or personal data.

\section*{Competing interests}
The author declares no competing interests.

% \section*{Author contributions}
% For double-anonymised review, author identifiers are redacted. Author contribution
% details will be supplied in the cover letter and restored in the accepted
% manuscript.


\begin{thebibliography}{34}
\providecommand{\natexlab}[1]{#1}
\providecommand{\url}[1]{\texttt{#1}}
\expandafter\ifx\csname urlstyle\endcsname\relax
  \providecommand{\doi}[1]{doi: #1}\else
  \providecommand{\doi}{doi: \begingroup \urlstyle{rm}\Url}\fi

\bibitem[Adams(1995)]{adams1995dandies}
James~Eli Adams.
\newblock \emph{Dandies and Desert Saints: Styles of Victorian Masculinity}.
\newblock Cornell University Press, Ithaca, NY, 1995.

\bibitem[Arora et~al.(2013)Arora, Ge, Halpern, Mimno, Moitra, Sontag, Wu, and
  Zhu]{arora2013practical}
Sanjeev Arora, Rong Ge, Yonatan Halpern, David Mimno, Ankur Moitra, David
  Sontag, Yichen Wu, and Michael Zhu.
\newblock {A Practical Algorithm for Topic Modeling with Provable Guarantees}.
\newblock In \emph{Proceedings of the 30th International Conference on Machine
  Learning (ICML)}, pages 280--288, 2013.

\bibitem[Bamman et~al.(2014)Bamman, Underwood, and Smith]{bamman2014bayesian}
David Bamman, Ted Underwood, and Noah~A. Smith.
\newblock {A Bayesian Mixed Effects Model of Literary Character}.
\newblock In \emph{Proceedings of the 52nd Annual Meeting of the Association
  for Computational Linguistics (ACL)}, pages 370--379, 2014.

\bibitem[Bederman(1995)]{bederman1995manliness}
Gail Bederman.
\newblock \emph{{Manliness and Civilization: A Cultural History of Gender and
  Race in the United States, 1880--1917}}.
\newblock University of Chicago Press, Chicago, 1995.

\bibitem[Bischof and Airoldi(2012)]{bischof2012summarizing}
Jonathan~M. Bischof and Edoardo~M. Airoldi.
\newblock {Summarizing Topical Content with Word Frequency and Exclusivity}.
\newblock In \emph{Proceedings of the 29th International Conference on Machine
  Learning (ICML)}, 2012.

\bibitem[Blei et~al.(2003)Blei, Ng, and Jordan]{blei2003lda}
David~M. Blei, Andrew~Y. Ng, and Michael~I. Jordan.
\newblock Latent {D}irichlet allocation.
\newblock \emph{Journal of Machine Learning Research}, 3:\penalty0 993--1022,
  2003.

\bibitem[Buurma(2015)]{buurma2015fictionality}
Rachel~Sagner Buurma.
\newblock The fictionality of topic modeling: Machine reading {A}nthony
  {T}rollope's {B}arsetshire series.
\newblock \emph{Big Data \& Society}, 2\penalty0 (2):\penalty0 1--6, 2015.
\newblock \doi{10.1177/2053951715610591}.

\bibitem[Chang et~al.(2009)Chang, Gerrish, Wang, Boyd-Graber, and
  Blei]{chang2009reading}
Jonathan Chang, Sean Gerrish, Chong Wang, Jordan~L. Boyd-Graber, and David~M.
  Blei.
\newblock Reading tea leaves: How humans interpret topic models.
\newblock In \emph{Advances in Neural Information Processing Systems 22}, 2009.

\bibitem[Connell(1995)]{connell1995masculinities}
R.~W. Connell.
\newblock \emph{Masculinities}.
\newblock University of California Press, Berkeley, CA, 1995.

\bibitem[Connell and Messerschmidt(2005)]{connell2005hegemonic}
R.~W. Connell and James~W. Messerschmidt.
\newblock Hegemonic masculinity: Rethinking the concept.
\newblock \emph{Gender \& Society}, 19\penalty0 (6):\penalty0 829--859, 2005.
\newblock \doi{10.1177/0891243205278639}.

\bibitem[Da(2019)]{da2019computational}
Nan~Z. Da.
\newblock The computational case against computational literary studies.
\newblock \emph{Critical Inquiry}, 45\penalty0 (3):\penalty0 601--639, 2019.
\newblock \doi{10.1086/702594}.

\bibitem[Davidoff and Hall(1987)]{davidoff1987family}
Leonore Davidoff and Catherine Hall.
\newblock \emph{Family Fortunes: Men and Women of the English Middle Class,
  1780--1850}.
\newblock University of Chicago Press, Chicago, IL, 1987.

\bibitem[Denny and Spirling(2018)]{denny2018text}
Matthew~J. Denny and Arthur Spirling.
\newblock {Text Preprocessing for Unsupervised Learning: Why It Matters, When
  It Misleads, and What to Do about It}.
\newblock \emph{Political Analysis}, 26\penalty0 (2):\penalty0 168--189, 2018.

\bibitem[Garg et~al.(2018)Garg, Schiebinger, Jurafsky, and Zou]{garg2018word}
Nikhil Garg, Londa Schiebinger, Dan Jurafsky, and James Zou.
\newblock Word embeddings quantify 100 years of gender and ethnic stereotypes.
\newblock \emph{Proceedings of the National Academy of Sciences}, 115\penalty0
  (16):\penalty0 E3635--E3644, 2018.
\newblock \doi{10.1073/pnas.1720347115}.

\bibitem[Girouard(1981)]{girouard1981camelot}
Mark Girouard.
\newblock \emph{{The Return to Camelot: Chivalry and the English Gentleman}}.
\newblock Yale University Press, New Haven, 1981.

\bibitem[Goldstone and Underwood(2014)]{goldstone2014quiet}
Andrew Goldstone and Ted Underwood.
\newblock The quiet transformations of literary studies: What thirteen thousand
  scholars could tell us.
\newblock \emph{New Literary History}, 45\penalty0 (3):\penalty0 359--384,
  2014.
\newblock \doi{10.1353/nlh.2014.0025}.

\bibitem[Hoyle et~al.(2021)Hoyle, Goel, Hian-Cheong, Peskov, Boyd-Graber, and
  Resnik]{hoyle2021automated}
Alexander Hoyle, Pranav Goel, Andrew Hian-Cheong, Denis Peskov, Jordan
  Boyd-Graber, and Philip Resnik.
\newblock Is automated topic model evaluation broken? the incoherence of
  coherence.
\newblock In \emph{Advances in Neural Information Processing Systems 34
  (NeurIPS 2021)}, 2021.

\bibitem[Jockers(2013)]{jockers2013macroanalysis}
Matthew~L. Jockers.
\newblock \emph{Macroanalysis: Digital Methods and Literary History}.
\newblock University of Illinois Press, Urbana, IL, 2013.

\bibitem[Jockers and Mimno(2013)]{jockers2013significant}
Matthew~L. Jockers and David Mimno.
\newblock Significant themes in 19th-century literature.
\newblock \emph{Poetics}, 41\penalty0 (6):\penalty0 750--769, 2013.
\newblock \doi{10.1016/j.poetic.2013.08.005}.

\bibitem[Kimmel(1996)]{kimmel1996manhood}
Michael~S. Kimmel.
\newblock \emph{{Manhood in America: A Cultural History}}.
\newblock Free Press, New York, 1996.

\bibitem[Mallett(2015)]{mallett2015victorian}
Phillip Mallett, editor.
\newblock \emph{The Victorian Novel and Masculinity}.
\newblock Palgrave Macmillan, London, UK, 2015.
\newblock \doi{10.1057/9781137491541}.

\bibitem[Mimno et~al.(2011)Mimno, Wallach, Talley, Leenders, and
  McCallum]{mimno2011optimizing}
David Mimno, Hanna~M. Wallach, Edmund Talley, Miriam Leenders, and Andrew
  McCallum.
\newblock Optimizing semantic coherence in topic models.
\newblock In \emph{Proceedings of the 2011 Conference on Empirical Methods in
  Natural Language Processing}, pages 262--272, Edinburgh, UK, 2011.

\bibitem[Mohr and Bogdanov(2013)]{mohr2013introduction}
John~W. Mohr and Petko Bogdanov.
\newblock {Introduction---Topic models: What they are and why they matter}.
\newblock \emph{Poetics}, 41\penalty0 (6):\penalty0 545--569, 2013.
\newblock \doi{10.1016/j.poetic.2013.10.001}.

\bibitem[Moretti(2013)]{moretti2013distant}
Franco Moretti.
\newblock \emph{Distant Reading}.
\newblock Verso, London, 2013.

\bibitem[Piper(2016)]{piper2016txtlab}
Andrew Piper.
\newblock txt{LAB} {N}ovel450: A multilingual data set of novels, 2016.
\newblock Dataset.

\bibitem[Roberts et~al.(2014)Roberts, Stewart, Tingley, Lucas, Leder-Luis,
  Gadarian, Albertson, and Rand]{roberts2014structural}
Margaret~E. Roberts, Brandon~M. Stewart, Dustin Tingley, Christopher Lucas,
  Jetson Leder-Luis, Shana~Kushner Gadarian, Bethany Albertson, and David~G.
  Rand.
\newblock Structural topic models for open-ended survey responses.
\newblock \emph{American Journal of Political Science}, 58\penalty0
  (4):\penalty0 1064--1082, 2014.
\newblock \doi{10.1111/ajps.12103}.

\bibitem[Roberts et~al.(2016)Roberts, Stewart, and Airoldi]{roberts2016model}
Margaret~E. Roberts, Brandon~M. Stewart, and Edoardo~M. Airoldi.
\newblock {A Model of Text for Experimentation in the Social Sciences}.
\newblock \emph{Journal of the American Statistical Association}, 111\penalty0
  (515):\penalty0 988--1003, 2016.
\newblock \doi{10.1080/01621459.2016.1141684}.

\bibitem[Roberts et~al.(2019)Roberts, Stewart, and Tingley]{roberts2019stm}
Margaret~E. Roberts, Brandon~M. Stewart, and Dustin Tingley.
\newblock {stm}: An {R} package for structural topic models.
\newblock \emph{Journal of Statistical Software}, 91\penalty0 (2):\penalty0
  1--40, 2019.
\newblock \doi{10.18637/jss.v091.i02}.

\bibitem[Sap et~al.(2017)Sap, Prasettio, Holtzman, Rashkin, and
  Choi]{sap2017connotation}
Maarten Sap, Marcella~Cindy Prasettio, Ari Holtzman, Hannah Rashkin, and Yejin
  Choi.
\newblock Connotation frames of power and agency in modern films.
\newblock In \emph{Proceedings of the 2017 Conference on Empirical Methods in
  Natural Language Processing (EMNLP)}, pages 2329--2334, 2017.

\bibitem[Solinger(2012)]{solinger2012becoming}
Jason~D. Solinger.
\newblock \emph{Becoming the Gentleman: British Literature and the Invention of
  Modern Masculinity, 1660--1815}.
\newblock Palgrave Macmillan, New York, NY, 2012.
\newblock \doi{10.1057/9780230391840}.

\bibitem[Sussman(1995)]{sussman1995victorian}
Herbert Sussman.
\newblock \emph{Victorian Masculinities: Manhood and Masculine Poetics in Early
  Victorian Literature and Art}.
\newblock Cambridge University Press, Cambridge, UK, 1995.

\bibitem[Tangherlini and Leonard(2013)]{tangherlini2013trawling}
Timothy~R. Tangherlini and Peter Leonard.
\newblock Trawling in the sea of the great unread: Sub-corpus topic modeling
  and humanities research.
\newblock \emph{Poetics}, 41\penalty0 (6), 2013.

\bibitem[Tosh(1999)]{tosh1999mans}
John Tosh.
\newblock \emph{A Man's Place: Masculinity and the Middle-Class Home in
  Victorian England}.
\newblock Yale University Press, New Haven, CT, 1999.

\bibitem[Underwood et~al.(2018)Underwood, Bamman, and
  Lee]{underwood2018transformation}
Ted Underwood, David Bamman, and Sabrina Lee.
\newblock The transformation of gender in {E}nglish-language fiction.
\newblock \emph{Journal of Cultural Analytics}, 2018.
\newblock \doi{10.22148/16.019}.

\bibitem[Vance(1985)]{vance1985sinews}
Norman Vance.
\newblock \emph{{The Sinews of the Spirit: The Ideal of Christian Manliness in
  Victorian Literature and Religious Thought}}.
\newblock Cambridge University Press, Cambridge, 1985.

\bibitem[Wallach et~al.(2009)Wallach, Murray, Salakhutdinov, and
  Mimno]{wallach2009evaluation}
Hanna~M. Wallach, Iain Murray, Ruslan Salakhutdinov, and David Mimno.
\newblock {Evaluation Methods for Topic Models}.
\newblock In \emph{Proceedings of the 26th Annual International Conference on
  Machine Learning (ICML)}, pages 1105--1112, 2009.

\end{thebibliography}
\end{document}